%% file: main.tex
\documentclass{article}

\usepackage[preprint,nonatbib]{./bib_bst_sty/neurips_2024}

\usepackage[utf8]{inputenc} 
\usepackage[T1]{fontenc}    
\usepackage{hyperref}       
\usepackage{url}            
\usepackage{booktabs}       
\usepackage{amsfonts}       
\usepackage{nicefrac}       
\usepackage{microtype}      
\usepackage{xcolor}         
\usepackage{graphicx}
\usepackage{enumitem}

\input{Definitions}

\title{Wavelet-Based Image Tokenizer for Vision Transformers}

\author{
  Zhenhai Zhu \\
  Google DeepMind \\
  \texttt{zhenhai@google.com}
  \\
  \And
  Radu Soricut \\
  Google DeepMind \\
  \texttt{rsoricut@google.com}
}

\begin{document}
\maketitle
\input{0_abstract}
\input{1_introduction}

\input{2_related_work}

\input{3_wavelet_tutorial}

\input{4_image_tokenizer}

\input{5_results}

\input{6_conclusions}

\newpage
\bibliography{./bib_bst_sty/attention, ./bib_bst_sty/nlp, ./bib_bst_sty/numerics, ./bib_bst_sty/vision} 
\bibliographystyle{plain}

\end{document}

%% file: 0_abstract.tex
\begin{abstract} 
Non-overlapping patch-wise convolution is the default image tokenizer for all state-of-the-art vision Transformer (ViT) models. Even though many ViT variants have been proposed to improve its efficiency and accuracy, little research on improving the image tokenizer itself has been reported in the literature. 
In this paper, we propose a new image tokenizer based on wavelet transformation. We show that ViT models with the new tokenizer achieve both higher training throughput and better top-1 precision for the ImageNet validation set. 
We present a theoretical analysis on why the proposed tokenizer improves the training throughput without any change to ViT model architecture. Our analysis suggests that the new tokenizer can effectively handle high-resolution images and is naturally resistant to adversarial attack. Furthermore, the proposed image tokenizer offers a fresh perspective on important new research directions for ViT-based model design, such as image tokens on a non-uniform grid for image understanding.
\end{abstract} 

%% file: 1_introduction.tex
\section{Introduction}
\label{sec:intro}
Natural languages inherently use a discrete set of characters or bytes in a fixed alphabet. A sequence of bytes can be directly treated as the input to a Transformer model~\cite{attention_vaswani2017}, as done in ByT5~\cite{ByT5_Xue2021}. However, the latent representations for such byte sequences may not carry enough information, and the sequence length could be unnecessarily long. A more common alternative is segmenting words into subword tokens either explicitly~\cite{SentencePiece_Kudo2018,WordPiece_Song2020} or implicitly~\cite{Charformer_Tay2021}. The explicit subword tokenizer has been a default component in many state-of-the-art (SOTA) large language models such as T5~\cite{t5_Raffel2019}, GPT3~\cite{GPT3_Brown2020}, ULM~\cite{Tay2022UL2UL}, PaLM ~\cite{PaLM_Chowdhery2022} and Chinchilla~\cite{chinchi_Hoffmann2022}.

Image pixels are also discrete since the RGB channels are represented by 8-bit unsigned integers. This means that the raw pixel vocabulary size is $2^{24}$, too large to be practical. In addition, the sequence length for an image with even a moderate resolution of $256 \times 256$ is too long. 
A commonly used image tokenizer for the vision Transformer (ViT)~\cite{VIT2020} is the non-overlapping patch-wise convolution. Despite many variants of ViT \cite{halonet2021,touvron2022deit3,liu2022swinv2,beyer2022flexivit,lee2022pix2struct}, this patch-convolution based tokenizer is still the default choice for SOTA ViT models \cite{steiner2021augreg,Zhai2021ScalingVT,NaViT_Dehghani2023}.
Since ViT models are the image encoders in the largest vision-language models such as CLIP~\cite{clip_Radford2021}, ALIGN~\cite{Align_Jia2021}, CoCa~\cite{CoCa_Yu2022} and PaLI~\cite{PaLI_chen2022}, the patch-convolution based image tokenizer becomes a default component in these vision-language models. 

An image tokenizer performs two main tasks: 1) it partitions a given image into non-overlapping patches so that the final token count is considerably smaller than the image pixel count; 2) it maps the RGB channels for the pixels in each patch into a token embedding vector. This converts a set of image patches to an input token sequence. The image tokenizer inevitably removes the relative position information among the pixels associated with each image token. A good image tokenizer must therefore strike a balance between the loss in position information and the savings in compute, both due to token-count reduction. At the same time, different vision tasks have different requirements for these two conflicting aspects. For datasets with high-resolution images~\cite{InfographicVQA_Mathew2021, lee2022pix2struct}, token count can easily become the bottleneck due to the quadratic complexity of the standard Transformer attention~\cite{SurveyEfficientTransformer}. For vision tasks like semantic segmentation~\cite{cityscape_Cordts2016, ADE20k_Zhou2016} and object detection~\cite{Pascal_Everingham2010, Kitti_Geiger2012}, the relative position or spatial information in general plays a crucial role. 
We believe that a good tokenizer should effectively compress the redundant information in patch pixels. This allows the flexibility to accommodate the two conflicting requirements, and has motivated us to look into the image compression field for inspiration.

There are many well-established and well-understood algorithms in the image compression field~\cite{imageProcessing_book}.
The central goal in image compression is to remove redundant information in pixel space such that the reconstructed images are visually close to the original images. The prevailing algorithm used in current industry standard JPEG2000~\cite{jpeg2000_book} is the wavelet transformation~\cite{wavelet_book_Daubechies,wavelet_book_Strang}. 

In this paper, we draw inspiration from the crucial insight in JPEG2000: most high-frequency signals revealed by the wavelet transformation can be safely truncated with minimal human perception difference. This is where the bulk of the image compression occurs. In short, we propose to replace the patch-convolution--based image tokenizer in the standard ViT architecture~\cite{VIT2020} with a wavelets-based image tokenizer. 

Our main \textbf{contributions} are:
\begin{enumerate}[leftmargin=4mm, itemsep=1mm, partopsep=0pt,parsep=0pt]
    \item 
    We introduce a new concept called pixel-space token embedding and show how to use wavelet transformation to compute it (Section \ref{subsec:pixel-space-embeddings}). 
    The pixel-space token embedding is naturally resistant to adversarial attacks and its sparsity can be used to produce image tokens on a non-uniform grid.
    \item 
    We present a theoretical analysis on how to use a block sparse projection to map truncated pixel-space token embeddings to semantic token embeddings with reduced dimensions (Section \ref{subsec:semantic-token-embeddings}).  This is important for training ViT models on datasets with images at higher resolutions.
    \item 
    The wavelet-based tokenizer offers an elegant solution to alleviating the computational impact of the two quadratic terms in Transformer layer op counts (Section \ref{subsec:op-count}). To the best of our knowledge, efficient Transformer research primarily focuses on the quadratic term $O(T^2)$, where $T$ is token count (sequence length). For some tasks and datasets, however, the op count is actually dominated by the other quadratic term $O(H^2)$, where $H$ is embedding size. Our new image tokenizer provides an efficient handle on this. 
    \item 
    We use empirical evidence to show that ViT models with the proposed wavelet-based tokenizer achieve higher training throughput and better top-1 precision (on the ImageNet val set, see Section~\ref{sec:results}), while reducing the overall model size. 
\end{enumerate}

%% file: 2_related_work.tex
\section{Related Work}
\label{sec:related_work}
The information redundancy in images has been previously exploited in the design of ViT variants. For instance, less informative patch-wise image tokens are omitted in later-stage of an adaptive ViT~\cite{AViT_Yin2021} or merged into tokens with larger but lower-resolution patches~\cite{Ronen2023VisionTW}. These approaches are less straightforward compared to ours, as they involve modification of ViT model, loss function and training recipe. Our wavelets-base tokenizer  offers a more principled framework to achieve the same goal, and it comes with a drop-in replacement of only the convolution-based tokenizer and no further changes. We will come back to this in section \ref{subsec:pixel-space-embeddings}.

Wavelet transformation has been used as a lossless up-sampling and down-sampling technique in ViT~\cite{Wave-ViT_Yao2022}. This approach is tangent to the image tokenizer proposed and studied in this paper.

JPEG-inspired algorithms have been proposed for image generation~\cite{DCT-image_Nash2021}, where the discrete cosine transformation (DCT) coefficients are directly generated by an autoregressive Transformer decoder. The sequence length is still a severe limiting factor, to the extent to which only low-resolution images can be generated. In our initial experiments, we used DCT as a building block for the image tokenizer, but the results were inferior to those obtained with wavelet-based image tokenizers.

Earlier image generation work directly decodes pixel RGB channels and can only generate low-resolution images~\cite{GPT-image, ImageTransformer2018}. Current SOTA text-to-image autoregressive generation models such as DALL-E~\cite{Dall-E_Ramesh2021} and others~\cite{ViT-VQGAN_Yu2022} all use a two-stage approach. The first stage trains an image tokenizer and detokenizer with a codebook of discrete visual tokens. The second stage trains a Transformer model that generates the final images from text prompts. 
The backbone in the tokenizers is either the discrete variational autoencoder (dVAE)~\cite{VAE_Oord2017} in DALL-E~\cite{Dall-E_Ramesh2021} or ViT-based VQGAN~\cite{Esser2020TamingTF} (ViT-VQGAN) in \cite{ViT-VQGAN_Yu2022}. They also often involve vector quantization in order to generate discrete visual tokens. Standard ViT models do not need a codebook with discrete tokens. Hence the tokenizers in~\cite{Dall-E_Ramesh2021, Esser2020TamingTF, ViT-VQGAN_Yu2022} are outside the scope of this paper. 
However, the convolutional neural network (CNN) based image encoder used by VQGAN in~\cite{Esser2020TamingTF} or dVAE in DALL-E~\cite{Dall-E_Ramesh2021} is related to our work because it has been recognized that the lower-level convolutional kernels in CNNs are very similar to Gabor wavelet filters~\cite{Luan2017GaborCN, Wang2023LearnableGK}. 
This area requires further investigation into connecting these approaches.

%% file: 3_wavelet_tutorial.tex
\section{A Brief Introduction to Wavelet-Based Image Compression}
\label{sec:wavelet_tutorial}

We first describe the discrete wavelet transformation in terms of convolution kernels. 
We refer the readers to excellent books such as \cite{wavelet_book_Daubechies,wavelet_book_Strang} for a comprehensive review of the theory on wavelet transformation.

A color image with resolution $N \times N$ and RGB channels can be represented by a tensor $P \in R^{N \times N \times 3}$. It is common practice in image compression to convert the RGB format to the luminance-chrominance format via a fixed and invertable linear projection~\cite{imageProcessing_book}. This is represented by another tensor $[Y, C_b, C_r] \in R^{N \times N \times 3}$ where $Y \in R^{N \times N}$ is the luminance matrix, and $C_b, C_r \in R^{N \times N}$ are the chrominance-blue and chrominance-red matrix, respectively. It is known that human visual perception is more sensitive to the brightness or luminance and less sensitive to the color or chrominance~\cite{imageProcessing_book}. Hence it is common practice to reduce the original image resolution for $C_b$ and $C_r$ by a factor of $2 \times 2$ before the wavelet transformation is applied. This reduces the overall pixel count by a factor of 2. 

There are many families of wavelets. Here we focus on two families that are effective for the purpose of image tokenizer: Daubechies (db) and Coiflets (coif). One can choose different orders within each family. For instance, the lowest-order one-dimensional (1D) kernels in db and coif families are, respectively, the db1 (also called Haar) kernels:
\begin{equation}
  K_l = 
  \left[
  \begin{array}{cc}
       \frac{1}{\sqrt{2}} & \frac{1}{\sqrt{2}}
    \end{array}
  \right], \;\;
  K_h = 
  \left[
  \begin{array}{cc}
       \frac{1}{\sqrt{2}} & -\frac{1}{\sqrt{2}}
    \end{array}
  \right]
\label{eq:1d-haar}
\end{equation}
and the coif1 kernels:
\begin{equation}
  \left[
  \begin{array}{c}
       K_l \\ K_h
    \end{array}
  \right]
=
  \left[
  \begin{array}{cccccc}
    -0.016 & -0.073 & 0.385 & 0.853 & 0.338 & -0.073 \\
    0.073 & 0.338 & -0.853 & 0.385 & 0.073 & -0.016
  \end{array}
  \right].
\label{eq:1d-coif1}
\end{equation}
The two-dimensional (2D) kernels are derived from 1D kernels as 
\begin{equation}
  K_{ll} = K_l^T K_l, \;\;\;
  K_{lh} = K_{l}^T K_{h}, \;\;\;
  K_{hl} = K_{h}^T K_{l}, \;\;\;
  K_{hh} = K_{h}^T K_{h}.
\label{eq:2d-kernel}
\end{equation}
The convolution with these four 2D kernels is then performed on $Y$, $C_b$ and $C_r$ separately. This is also known as the group convolution. For instance, convolution on matrix $Y$ results in
\begin{eqnarray}
  Y^{(1)} &=& \mathbf{conv_{2d}}(K_{ll})(Y^{(0)}), \;\;\;
  Y_{h}^{(1)} = \mathbf{conv_{2d}}(K_{lh})(Y^{(0)})
\label{eq:A_r} \\
  Y_v^{(1)} &=& \mathbf{conv_{2d}}(K_{hl})(Y^{(0)}), \;\;\;
  Y_d^{(1)} = \mathbf{conv_{2d}}(K_{hh})(Y^{(0)})
\label{eq:Y_d}
\end{eqnarray}
where $Y^{(0)}=Y$ and $\mathbf{conv_{2d}}(M)$ is the 2D convolution with kernel $M$ and a fixed stride $(2,2)$. The same convolution in equations (\ref{eq:A_r})-(\ref{eq:Y_d}) can be recursively performed on matrix $Y^{(k)}$ for $k=1,2...$. The stride $(2,2)$ dictates that each recursion step reduces the image resolution by a factor of $2 \times 2$. 
Matrices $Y^{(k)}$, $Y_h^{(k)}$, $Y_v^{(k)}$ and $Y_d^{(k)}$ can be gathered into a single matrix. For instance, a simple level-2 wavelet transformation on matrix $Y$ results in 
\begin{equation}
    \left[\begin{array}{@{}c|c@{}}
    \begin{array}{c|c}
       Y^{(2)} & Y_h^{(2} \\
        \hline
       Y_v^{(2)} & Y_d^{(2)}
    \end{array}
    & Y_h^{(1)} \\ \\
    \hline
       Y_v^{(1)} & Y_d^{(1)}
    \end{array}\right].
\label{eq:gathered_Y}
\end{equation}
The procedure in equations (\ref{eq:A_r})-(\ref{eq:gathered_Y}) is also applied to matrices $C_b$ and $C_r$, and we obtain the same matrix structure shown in equation (\ref{eq:gathered_Y}) for $C_b$ and $C_r$.

\begin{figure}[h]
    \includegraphics[width=\linewidth]{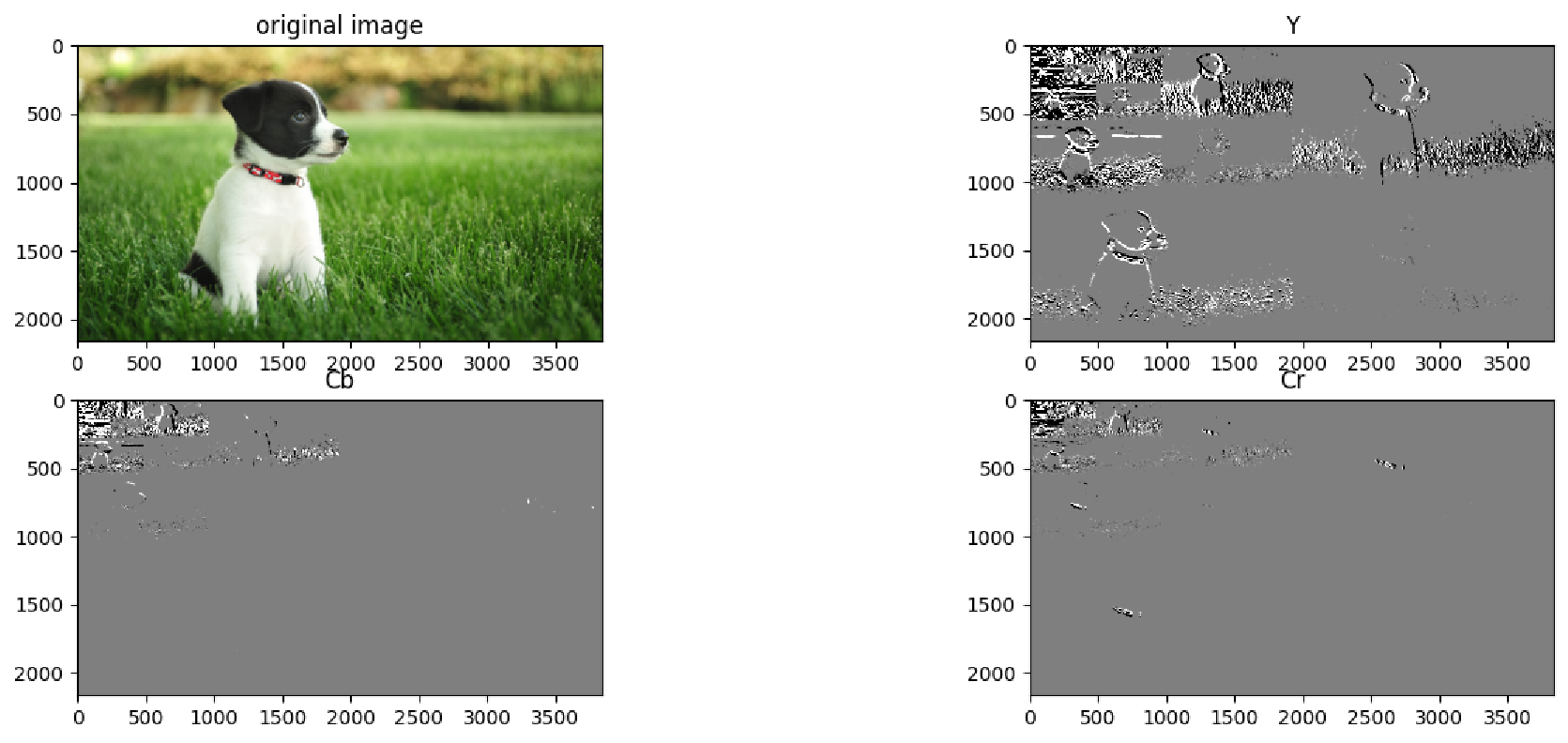}
    \centering
    \caption{Original image and its wavelet coefficients for $YC_bC_r$ channels, where black/white/grey dots correspond to strongly positive / strongly negative / near-to-zero coefficients, respectively.}
    \label{fig:puppy_wt}
\end{figure}

Figure \ref{fig:puppy_wt} shows a level-10 wavelet transformation using the db1 kernels of equation (\ref{eq:1d-haar}) on an image with $YC_bC_r$ channels. Notice that majority of the wavelet coefficients are nearly zero. This is particularly true for channel-$C_b$ and channel-$C_r$. This distribution is typical of a natural image and it gives us plenty of room to safely sparsify or zero out those near-to-zero wavelet coefficients with minimal quality loss in the reconstructed image. All non-zero coefficients are quantized and then compressed further with entropy coding algorithms. We refer interested readers to \cite{jpeg2000_book} for details on this step.

We can easily reverse the steps in equations (\ref{eq:A_r})-(\ref{eq:gathered_Y}) and reconstruct the original image. The main difference is that we need to use transposed convolution in equations (\ref{eq:A_r})-(\ref{eq:Y_d}).

%% file: 4_image_tokenizer.tex
\section{Image Tokenizer}
\label{sec:tokenizer}
In this section, we depart from the image compression concepts behind the standard JPEG2000 algorithm and focus on designing an effective image tokenizer. The common goal is to remove redundant information in an image, hence wavelet transformation is a shared component. However, the end goals are different. In an image compression algorithm, each step  must be reversible such that the original image can be reconstructed from JPEG strings with minimal visual-perception quality loss. On the other hand, an image tokenizer is designed to capture semantically meaningful information. The generated token embeddings serve as the starting point for the subsequent processing performed by ViT models, namely, the "echo chambers" effect in attention layers~\cite{Glow2021} and the key-value pair memories in feedforward layers~\cite{Geva2020TransformerFL}. 

\subsection{Pixel-Space Token Embedding}
\label{subsec:pixel-space-embeddings}
Same as in JPEG2000, we adopt the RGB-to-$YC_bC_r$ linear transformation and downsample $C_b$ and $C_r$ channels by a factor $2 \times 2$. Hence for an image with resolution $N \times N$, the input to the wavelet transformation is $YC_bC_r$ channels with shape $Y \in R^{N \times N}$ and $C_b, C_r \in R^{\frac{N}{2} \times \frac{N}{2}}$. 

Unlike Fourier or discrete cosine transformation, wavelet transformation uses localized kernels, see equations (\ref{eq:1d-haar}) and (\ref{eq:1d-coif1}). Therefore, each wavelet coefficient is related only to a local pixel region. 
We can illustrate this with a concrete example. 
Given a color image with resolution $128 \times 128$, we follow the procedure in equations (\ref{eq:A_r})-(\ref{eq:Y_d}) and perform a level-2 wavelet transformation on $YC_bC_r$ channels. We can re-write the resulting matrix in equation (\ref{eq:gathered_Y}) and its counter parts for $C_b$ and $C_r$ as:
\begin{eqnarray}
  P^{(2)} &=& 
  \left[
  \begin{array}{ccc}
    Y^{(2)} & C_b^{(2)} & C_r^{(2)}
  \end{array}
  \right]
\label{eq:coarse_image_P2} \\
  D^{(2)} &=& 
  \left[
  \begin{array}{ccccccccc}
    Y_h^{(2)} & Y_v^{(2)} & Y_d^{(2)} &
    C_{bh}^{(2)} & C_{bv}^{(2)} & C_{bd}^{(2)} &
    C_{rh}^{(2)} & C_{rv}^{(2)} & C_{rd}^{(2)}
  \end{array}
  \right]
\label{eq:coarse_image_D2} \\
  D^{(1)} &=& \left[
  \begin{array}{ccc}
    Y_h^{(1)} & Y_v^{(1)} & Y_d^{(1)} 
    \end{array}
  \right]
\label{eq:coarse_image_D1}
\end{eqnarray}
where $P^{(2)} \in R^{32 \times 32 \times 3}$, $D^{(2)} \in R^{32 \times 32 \times 9}$ and $D^{(1)} \in R^{64 \times 64 \times 3}$.
Note that we have skipped the convolution on $C_b$ and $C_r$ at resolution $128 \times 128$ since they are already downsampled to $64 \times 64$. This is why there is no $C_b$ or $C_r$ component in equation (\ref{eq:coarse_image_D1}). Note also that $P^{(2)}$ is just a coarsened image with $YC_bC_r$ channels at resolution $32 \times 32$. We can reshape tensor $D^{(1)}$ such that $D^{(1)} \in R^{32 \times 32 \times 12}$ and concatenate $P^{(2)}$, $D^{(2)}$ and $D^{(1)}$ along their last dimension into one tensor as
\begin{equation}
  W = 
  \left[
  \begin{array}{ccc}
    P^{(2)} & D^{(2)} & D^{(1)}
    \end{array}
  \right]
  =
  \left[
  \begin{array}{ccc}
    W^{(2)} & W^{(1)}
    \end{array}
  \right]
\label{eq:stacked_w_level2}
\end{equation}
where $W \in R^{32 \times 32 \times 24}$, $W^{(2)}=[P^{(2)}\;\; D^{(2)}] \in R^{32 \times 32 \times 12}$ and $W^{(1)}=D^{(1)}$.
More generally, a level-$L$ wavelet transformation on an image with resolution $N \times N$ results in a set of tensors concatenated along their last dimension as
\begin{equation}
  W = 
  \left[
  \begin{array}{cccc}
    W^{(L)} & W^{(L-1)} & \hdots & W^{(1)}
    \end{array}
  \right]
\label{eq:stacked_w}
\end{equation}
where $W^{(k)} \in R^{\Tilde{N} \times \Tilde{N} \times C_k}$, $\Tilde{N} = \frac{N}{2^{L}}$ and
\begin{equation}    
  C_k = 
  \left\{
  \begin{array}{ll}
    12 & k=L 
    \\
    9 \times 4^{L-k} & k=2,...L-1
    \\
    3 \times 4^{L-1} & k=1
  \end{array}
  \right.
\label{eq:coef_count_level_k}
\end{equation}
The number of entries in vector $W(i,j,:)$ in equation (\ref{eq:stacked_w}) is
\begin{equation}
  C = \sum_{k=1}^{L} C_k = \frac{3}{2} 4^L.
\label{eq:coef_count}
\end{equation}
We treat the length-$C$ vector $W(i,j,:)$ as the pixel-space embedding for pixel-$(i,j)$ in the coarse image with resolution $\tilde{N} \times \tilde{N}$. 
For the simple level-2 example in equation (\ref{eq:stacked_w_level2}), we only need vector $W(i,j,:)$ with its 24 entries to reconstruct the $4 \times 4$ original pixel region corresponding to pixel-$(i,j)$ in coarse image $P^{(2)}$. This can be done by following steps in equations (\ref{eq:A_r})-(\ref{eq:stacked_w_level2}) in reverse order.
The same procedure also applies to the general case in equation (\ref{eq:stacked_w}). This reconstruction is exact if there is no truncation, i.e. no small wavelet coefficients are zeroed out. As shown in Fig \ref{fig:puppy_wt}, truncation via thresholding can be done easily. JPEG2000 has demonstrated that the reconstruction error is small even if more than $90\%$ of the wavelet coefficients are zeroed out. Therefore, the pixel-space token embedding from this point on always refers to the post-truncation version of vector $W(i,j,:)$. 

A few remarks are in order: 
\begin{enumerate}[leftmargin=4mm, itemsep=1mm, partopsep=0pt,parsep=0pt]

    \item 
    Token count in pixel-space embedding tensor $W$ is $\tilde{N}^2 = \frac{N^2}{4^L}$. So level-$L$ is used as a hyper-parameter to achieve the trade-off between token count reduction and the loss of positional information, as discussed in Section~\ref{sec:intro}. 

    \item 
    Wavelet kernels are fixed, as shown in equations (\ref{eq:1d-haar}) and (\ref{eq:1d-coif1}). So any gradient-based adversarial examples~\cite{adversarial_Goodfellow} would fail since there is no gradient with respect to these kernels. In general, any small white-noise-like perturbation to the input images becomes high-frequency wavelet coefficients~\cite{Bayer2019AnIW, Tian2022MultistageID}, which are zeroed out in the sparsification step. Therefore, a wavelet-based tokenizer is naturally resistant to adversarial attacks on image classifiers (e.g., ~\cite{Sen2023AdversarialAO}).

    \item 
    Wavelet transformation exposes image regions with smooth RGB pixels. The sparsity in each pixel-space embedding vector correlates strongly with the information entropy in the corresponding pixel region. This can be used as a useful signal in the design and implementation of the image tokenizer. For instance, it can be used to guide a non-uniform image partition, and produce image tokens on a non-uniform grid\footnote{The non-uniform grid or mesh is a standard technique in numerical methods for partial differential equations~\cite{fd_pde_book, fdfv_pde_book}. It is an indispensable tool for handling large variance in variables such as electric potential or airflow pressure as a function of spatial coordinates. The RGB or $YC_bC_r$ intensities in natural images also vary considerably with pixel coordinates.}. This is in spirit similar to the goal in~\cite{AViT_Yin2021, Ronen2023VisionTW}, but wavelet transformation provides a more principled framework. This involves modification to the ViT model itself and hence is outside the scope of this paper.

\end{enumerate}

\subsection{Semantic Token Embedding}
\label{subsec:semantic-token-embeddings}
The goal of this next step is to map the pixel-space embeddings to semantically-meaningful token embeddings. This mapping can be a simple trainable linear projection
\begin{equation}
  E = W Q
\label{eq:out_proj}
\end{equation}
where $E \in R^{\tilde{N} \times \tilde{N} \times H}$ is the semantic token embedding tensor, $H$ is the target embedding size, $Q \in R^{C \times H}$ is the trainable projection matrix, and $W \in R^{\tilde{N} \times \tilde{N} \times C}$  is defined in equation (\ref{eq:stacked_w}).

Equation (\ref{eq:coef_count}) shows that pixel-space embedding size grows exponentially with the wavelet level $L$. However, each vector $W(i,j,:)$ in equation (\ref{eq:stacked_w}) is extremely sparse, with a predictable sparsity pattern.
For the level-10 wavelets shown in Figure \ref{fig:puppy_wt}, nearly all entries in $W^{(1)}$ are zeros, whereas $W^{(L)}$ has almost no zero entries. 
In general, the proportion of zeros in $W^{(k)}$ grows rapidly, maybe exponentially, as level index $k$ decreases. 
Our experiments (Tables~\ref{tab:res_256}-\ref{tab:res_512}) show that sparsifying 80\% of the entries in $W$ actually boosts the image classification top-1 precision. We believe the reason is that those small wavelet coefficients contain more noise than signal, and filtering them out serves as a form of regularization for model training.

We can use a block sparse structure in matrix $Q$ to exploit this sparsity pattern. The new projection is
\begin{equation}
E  = WQ = W
  \left[
  \begin{array}{cccc}
    Q^{(L)} & 0 & \hdots & 0 \\
    0 & Q^{(L-1)} & \hdots & 0 \\
    \vdots & \vdots & \ddots & 0 \\
    0 & 0 & 0 & Q^{(1)}
    \end{array}
  \right]
\label{eq:level_proj}
\end{equation}
where $Q^{(k)} \in R^{C_k \times H_k}$, $C_k$ is defined in equation (\ref{eq:coef_count_level_k}), and $H_k$ is the token embedding size assigned to level-$k$ under the constraint 
\begin{equation}
  \sum_k H_k = H.
\label{eq:stacked_H}
\end{equation}
In view of equation (\ref{eq:stacked_w}), we have
\begin{equation}
E = 
  \left[
  \begin{array}{cccc}
    E^{(L)} & E^{(L-1)} & \hdots & E^{(1)}
  \end{array}
  \right]
\label{eq:level_proj_stacked}
\end{equation}
where
\begin{equation}
 E^{(k)} = W^{(k)} Q^{(k)}.
\label{eq:level_proj_k}
\end{equation}
The key question is: what is a right value for $H_k$ given $C_k$ in equation (\ref{eq:coef_count_level_k})?
This obviously depends on the sparsity of vector $W^{(k)}(i,j,:)$. For higher wavelet levels like $k=L$ or $k=L-1$,  tensor $W^{(k)}$ is dense and hence we can simply set $H_k=C_k$. It is the much larger projection matrix like $Q^{(1)}$ that makes a significant difference to the parameter count and runtime. Fortunately, tensor $W^{(1)}$ is very sparse and this offers a good opportunity to find a $H_k$ much smaller than $C_k$.

To facilitate the exposition, we focus on the calculation of vector $E^{(k)}(i,j,:)$
\begin{equation}
 E^{(k)}(i,j,:) = e^T_{ij} = W^{(k)}(i,j,:) Q^{(k)} = w^T_{ij} Q^{(k)}
\label{eq:level_proj_k_ij}
\end{equation}
or equivalently
\begin{equation}
 e_{ij} = (Q^{(k)})^T w_{ij}
\label{eq:level_proj_k_ij_t}
\end{equation}
where $e_{ij} \in R^{H_k \times 1}$, $w_{ij} \in R^{C_k \times 1}$, and $(Q^{(k)})^T \in R^{H_k \times C_k}$. Note that the positions of nonzero entries in vector $w_{ij}$ are random, depending on the image being processed. 
We show in two steps that we could set $H_k$ to the expected numerical rank of matrix $Q^{(k)}$, which is much smaller than $C_k$ for small $k$. The first step can be formalized as the following proposition:
\paragraph{Proposition 1}
For a linear projection
\begin{equation}
  \vec{y} = A \vec{x}
  \label{eq:linear_proj}
\end{equation}
where $A \in R^{M \times N}$, $\vec{x} \in R^{N}$ and only a random subset of entries in $\vec{x}$ are nonzero. The probability of an entry in $\vec{x}$ being nonzero is $p$. Suppose we sample such a Bernoulli distribution to generate a random $\vec{x}$ and accumulate the resulting $\vec{y}$ into a matrix
\begin{equation}
  B = \left[\vec{y}_1, \vec{y}_2, ...\right],
  \label{eq:linear_proj_result}
\end{equation}
then the expected column rank of $B$ is $pN$.
\begin{proof}
Column rank of A is $N$, but only $pN$ of them are used to compute each $\vec{y}_i$. So the columns of matrix B can only span a space covered $pN$ columns of matrix $A$ in probability. This is the expected column rank of matrix $B$.
\end{proof}

Suppose we collect all embedding vectors $e_{ij}$ in equation (\ref{eq:level_proj_k_ij_t}) into a single matrix $S$. Let the total token count be $S_t$ and hence 
$S \in R^{H_k \times S_t}$. Note that $S_t$ could be a huge number since matrix $S$ includes every level-$k$ pixel-space embedding vector in tensor $W^{(k)}$ extracted from all images in the training dataset. So obviously $H_k \ll S_t$. The second step can be stated as the next proposition.
\paragraph{Proposition 2}
The expected numerical rank of matrix $S$ is $min(H_k, pC_k)$. 
\begin{proof}
In view of equation (\ref{eq:level_proj_k_ij_t}) and (\ref{eq:linear_proj}), we can see that $e_{ij}$ corresponds to $y$ and $w_{ij}$ corresponds to $x$. Likewise, matrix $S$ corresponds to $B$ in equation (\ref{eq:linear_proj_result}). As a direct result from Proposition-1, the expected column rank of matrix $S$ is $pC_k$. Hence $rank(S)=min(H_k, pC_k)$.
\end{proof}

Since we have the flexibility to set $H_k$ such that the information in $w_{ij}$ is preserved in $e_{ij}$ after the projection in equation (\ref{eq:level_proj_k_ij_t}), we can simply set $H_k=p C_k$.
Even though $C_k$ in equation (\ref{eq:coef_count_level_k}) can be large for small $k$ such as $k=1$, empirically we see that probability $p$ is very small for small $k$. This means that $H_k=p C_k$ is nearly constant for small $k$. For large $k$ such as $k=L$, we set $H_k=C_k$ since the pixel-space embeddings are fully dense or equivalently $p=1$. In view of equation (\ref{eq:stacked_H}), the semantic token embedding size $H$ is close to an affine function instead of an exponential function in wavelet level $L$. 

Consider an image with resolution $N \times N$, patch-convolution based tokenizer with patch size $p \times p$ generates $T=\frac{N^2}{p^2}$ tokens. Patch convolution becomes ineffective at extracting visual information when $p$ is too large, which is why $p=8,16,32$ are common choices. Hence the operation count for the standard ViT with patch-convolution tokenizer grows as $O(T^2)=O(\frac{N^4}{p^4})$. This becomes prohibitively expensive when $N>2048$. However, this high image resolution can be reduced to a user-specified resolution $\tilde{N}=\frac{N}{2^L}$ with a level-L wavelet-based tokenizer, where $L=log_2(\frac{N}{\tilde{N}})$. Since semantic token embedding size $H$ in equation (\ref{eq:stacked_H}) is affine in $L$, as discussed above, we have $H=O(log_2(\frac{N}{\tilde{N}}))$ which grows slowly with $N$. 
Therefore, wavelet-based tokenizer offers a much more efficient way to handle high image resolution due to 
the block sparse projection in equation (\ref{eq:level_proj}).

We could add more dense layers on top of the single linear layer in equation (\ref{eq:level_proj}) to potentially improve the token embedding quality. The increase in parameter count and training cost due to the added layers is $O(H^2)$. This is much smaller than $O(CH)$ in equation (\ref{eq:level_proj}) since the embedding dimension has been reduced from a large $C$ in equation (\ref{eq:coef_count}) to a small $H$ in equation (\ref{eq:stacked_H}).

\subsection{Operation Count For Transformer Layers}
\label{subsec:op-count}
A standard Transformer consists of many identical layers with the following hyper parameters: token embedding size $H$, model hidden state size which is typically the same $H$, attention query, key and value feature size $d_{model}$ which is typically also the same $H$, and feed-forward intermediate layer activation size $d_{ff}=m \times H$ (with integer $m$ typically in the range $[2, 10]$). 
As before, let $T$ be the token count. The breakdown of the operation count for each Transformer layer is
\begin{itemize}
    \item Attention: $2T^2 H + 4TH^2$
    \begin{itemize}
        \item Query/Key/Value multihead projections and output projection: $4 T H^2$
        \item Query Key inner-product and Attention Value multiplication: $2T^2 H$
    \end{itemize}
    \item Feedforward: $2m T H^2$
\end{itemize}
We focus our attention on the two quadratic terms: $2T^2 H$ and $(2m+4)T H^2$.
If we use a typical $m=4$, then the ratio between them is
\begin{equation}
  r = \frac{2T^2 H}{(2\times4+4)T H^2} = \frac{T}{6H}.
\label{eq:runtime_ratio}
\end{equation}
Much of the efficient Transformer research focuses on improving the $O(T^2)$ quadratic complexity in Transformer attention~\cite{SurveyEfficientTransformer, HTransformer1D_Zhu2021, RoutingTransformer, Performer2020, Linformer2020, Reformer2020, BigBird2020, ETC2020, Longformer2020}. This is important for tasks with long-context such as long-document summarization and long-video understanding and generation because $T \gg H$ and hence $r \gg 1$ in equation (\ref{eq:runtime_ratio}). However, the other quadratic term $O(H^2)$ dominates for many standard language or vision models and datasets. For example, $T=512, H=1024$ are used in a T5-large model~\cite{t5_Raffel2019}, and $T=256, H=4096$ are used in ViT image encoder in SOTA vision-language model PaLI~\cite{PaLI_chen2022}.
In such cases, $r \ll 1$ in equation (\ref{eq:runtime_ratio}) and hence it is more cost-effective to reduce $H$ using the wavelet-base image tokenizer with the block sparse projection from Section~\ref{subsec:semantic-token-embeddings}. In Section~\ref{sec:results} we confirm empirically that this is exactly what is observed for ViT models.

We should note that matrix-tiling and operation-fusion idea in the recently proposed Flash Attention~\cite{FlashAttention2_Dao2023, FlashAttention_Dao2022} has essentially removed the quadratic memory complexity for Transformer models. So we do not include memory usage in our analysis.
The discussion here only focus on the forward path in model training. The analysis for the backward gradient propagation path follows the same reasoning and the same conclusion applies.

%% file: 5_results.tex
\section{Experimental Results}
\label{sec:results}

\subsection{Setup}

\paragraph{Dataset} 
We use the ILSVRC-2012 ImageNet dataset with 1k classes and 1.3M images~\cite{russakovsky2014imagenet}. The image resolution has two settings: $256 \times 256$ and $512 \times 512$. 

\paragraph{ViT Models}
Our baseline model is the ViT-base model in~\cite{VIT2020} with the same model parameters: $H=d_{model}=768, d_{ff}=3072$, 12 Transformer layers, 12 attention heads. 
We treat $d_{model}$ and $H$ as two independent hyper parameters so we can check their separate effects on training throughput and top-1 precision.
All models are trained on TPUs v4~\cite{TPU_10lessone_2021, TPUv4_2023} with the same training hyper parameters reported in~\cite{VIT2020}, such as batch size, epochs, learning rate, weight decay and etc. 
We follow the data augmentation recipe in~\cite{steiner2021augreg} which uses the combination of Mixup~\cite{mixup_Zhang2017} and RandAugment~\cite{Randaugment_Cubuk2019}.
The results below clearly show that wavelet-based tokenizer does not interfere with these data augmentation tricks. 
We want to emphasize that ViT model architecture and training recipe are held unchanged as we do a drop-in replacement of patch-convolution image tokenizer with our wavelet-based tokenizer. This way, the improvement can directly be attributed to the wavelet tokenizer.

\paragraph{Patch-convolution tokenizer variants}
are controlled by patch size only. We use the notations patch/8, patch/16 and patch/32 in table \ref{tab:res_256} and table \ref{tab:res_512} to denote these variants. For instance, patch/8 means that a $8 \times 8$ pixel region corresponds to one ViT input token.

\paragraph{Wavelet-based tokenizer variants}
are controlled by two groups of parameters: 

1) The wavelet transformation parameters include wavelet type and level-$L$, as well as the percentage of wavelet coefficients to be zeroed out with thresholding. We use the notations db1-4 and db1-5 for level-4 and level-5 wavelet transformations, respectively, with the db1 kernels defined in equation (\ref{eq:1d-haar}). Similarly, coif1-3 and coif1-4 refer to level-3 and level-4 wavelet transformations, respectively, with the coif1 kernels defined in equation (\ref{eq:1d-coif1}). By default, $80\%$ of the wavelet coefficients of $W$ in equation (\ref{eq:stacked_w}) are zeroed out. This percentage is appropriate for image resolutions $256 \times 256$ and $512 \times 512$ (it should be higher for higher-resolution images).

2) The dense projection parameters include $H_k$ in equation (\ref{eq:stacked_H}) and number of additional dense layers. As discussed in section \ref{subsec:semantic-token-embeddings}, we set $H_k=C_k$ for $k=L,L-1$. 
We perform hyper parameter sweep for $H_1$ such that the top-1 precision is optimized under the constraint that $H$ in equation (\ref{eq:stacked_H}) is an integer multiple of 128. Two top performing choices ($H=256, 384$) and their impact are shown in Table~\ref{tab:embed_size}.
Our experiments show that one additional dense layer does not bring any gain in top-1 precision. This is likely due to the fact that $L$ is only up to 5 for image resolution $512 \times 512$. We believe more dense layers will be needed for resolution higher than $1024 \times 1024$.

\paragraph{Input sequence length}
It should be noted that each wavelet-based token corresponds to a pixel region with size $2^L \times 2^L$ in the original image. So the input sequence  generated from an image with resolution $256 \times 256$ by coif1-3 and patch/8 have the same length $(\frac{256}{8})^2=32^2=1024$. This sequence length is denoted as $tokens$ in all result tables.

\subsection{Tokenizer Comparison}
We run experiments comparing patch-convolution--based tokenizers (denoted as patch/$N$, where $N$ is the patch size) versus wavelet-based tokenizers (db and coif). We report results on images with resolutions $256 \times 256$ (Table \ref{tab:res_256}) and $512 \times 512$ (Table \ref{tab:res_512}). The training is done using TPUv4 accelerators, and throughput is measured in number of images per second per TPU core.
The results show that the wavelet-based tokenizers have both better top-1 precision and higher training throughput. Remarkably, wavelet-based tokenizers also reduce model parameter count significantly because $H$ is reduced, due to the block sparse projection in equation (\ref{eq:level_proj}).

\begin{table*}[ht]
\caption{Top-1 precision and throughput (plus stats) for ImageNet-1K (val set), 256x256 image resolution; patch/8,16 are patch-convolution tokenizers, coif1-3 and db1-4 are wavelet tokenizers.}
\begin{tabular}{l|c|c|ccccc}
tokenizer & prec@top-1 ($\uparrow$) & throughput ($\uparrow$) & params (M) ($\downarrow$) & tokens & $H$ & $d_{model}$ & $d_{ff}$\\ 
\hline
patch/8 & 82.21 & 99.8 & 86.76 & 1024 & 768 & 768 & 3072\\
coif1-3  & {\bf 83.43} {\tiny +1.22} & {\bf 116.5} {\tiny +17\%} & {\bf 44.05} {\tiny +49\%} & 1024 & 384 & 768 & 3072\\
\hline
patch/16 & 80.30 & 589  & 86.61  & 256 & 768 & 768 & 3072\\
db1-4   & {\bf 81.56} {\tiny +1.26} & {\bf 996} {\tiny +69\%} & {\bf 43.87} {\tiny +49\%} & 256 & 384 & 768 & 3072\\
\hline
\end{tabular}
\label{tab:res_256}
\end{table*}

\begin{table*}[ht]
\caption{Top-1 precision and throughput (plus stats) for ImageNet-1K (val set), 512x512 image resolution; patch/16,32 are patch-convolution tokenizers, coif1-4 and db1-5 are wavelet tokenizers.}
\begin{tabular}{l|c|c|ccccc}
tokenizer & prec@top-1 ($\uparrow$) & throughput ($\uparrow$) & params (M) ($\downarrow$) & tokens & $H$ & $d_{model}$ & $d_{ff}$\\ 
\hline
patch/16 & 82.86 & 99.8 & 87.2  & 1024 & 768 & 768 & 3072\\
coif1-4  & {\bf 83.14} {\tiny +0.28} & {\bf 102.1} {\tiny +2\%}& {\bf 44.31} {\tiny +49\%} & 1024 & 384 & 768 & 3072\\
\hline
patch/32 & 80.89 & 311  & 88.38  & 256 & 768 & 768 & 3072\\
db1-5   & {\bf 81.63} {\tiny +0.74}& {\bf 341} {\tiny +10\%} & {\bf 59.59} {\tiny +33\%} & 256 & 512 & 768 & 3072\\
\hline
\end{tabular}
\label{tab:res_512}
\end{table*}

\subsection{Effect of Token Embedding Sizes}
We report the results of using different $H$, $d_{ff}$ and $d_{model}$ in Table \ref{tab:embed_size}. 
Smaller $H$ allows us to make some useful trade-off. For instance, configuration $H=384, d_{model}=768$ means that the multihead projection size is down by $\frac{1}{2}$ and the feedforward layer size is down by a factor of $\frac{3}{4}$. This saving allows us to enlarge $d_{ff}$ from $4 \times H$ to up to $16 \times H$. This flexibility allows the trade-off between precision and training throughput, as shown in Table \ref{tab:embed_size}. 

\begin{table*}[h]
\caption{Study on the effect of different $H$, $d_{ff}$ and $d_{model}$ on Top-1 precision for ImageNet-1K (val set) and training throughput, 256x256 image resolution.}
\begin{tabular}{cc|c|c|cccc}
tokenizer & tokens & prec@top-1 ($\uparrow$) & throughput ($\uparrow$) & params (M) & $d_{model}$ & H & $d_{ff}$\\ 
\hline
db1-4 & 256 & 79.44 & 1422 & 14.21 & 256 & 256 & 3072 \\ 
db1-4 & 256 & 80.44 & 1101 & 29.52 & 768 & 256 & 3072 \\ 
db1-4 & 256 & 81.12 & 864 & 48.43 & 768 & 256 & 6144 \\ 
\hline
db1-4 & 256 & 81.22 & 1052 & 36.77 & 384 & 384 & 3072 \\ 
db1-4 & 256 & 81.56 & 996 & 43.87 & 768 & 384 & 3072 \\ 
db1-4 & 256 & 81.55 & 769 & 72.21 & 768 & 384 & 6144 \\ 
\hline
\end{tabular}
\label{tab:embed_size}
\end{table*}

%% file: 6_conclusions.tex
\section{Conclusions And Future Work}
\label{sec:conclusions}
We describe a new wavelet-based image tokenizer as a drop-in replacement for the patch-convolution based tokenizer used in the standard ViT models. We demonstrate that ViT models equipped with the new tokenizer achieve both higher training throughput and better top-1 precision for the ImageNet validation set. We also present a theoretical analysis on why the proposed tokenizer improves the training throughput, without any change to the ViT model architecture.  
Our analysis also indicates that the new tokenizer can effectively handle high-resolution images and is naturally resistant to adversarial attacks. 
Overall, our image tokenizer offers a fresh perspective on important new research directions for ViT-based model design, such as image tokens on a non-uniform grid and a new approach to modeling increased resolution in images.